*KEYNOTE LECTURE*

Akshar Bharati
Vineet Chaitanya
Rajeev Sangal
Language Technologies Research Centre
Indian Institute of Information Technology Hyderabad
{vc,sangal}@iiithyd.ernet.in

**OUTLINE**
 1. What is information revolution?
 2. What are the advantages of information in electronic form?
 3. What is hyper-text?
 4. Where are Indian languages in the information revolution?
   4.1 Digital content
   4.2 Machine translation
   4.3 Speech
   4.4 Optical character recognition
   4.5 Search and knowledge organization
   4.6 Other applications
 5. Relevant technologies
   5.1 Basic research
 6. Are computers affordable?
   6.1 Innovative solutions
   6.2 Sharing the use of computers
   6.3 Future cost projections
 7. What is the status of Indian language content in electronic form?
 8. How can we create electronic content in Indian languages?
 9. Can the collaborative method for creation of digital content work?
 10. What can I do? what are some of the concrete tasks that can
    10.1 Dictionary refinement
    10.2 Dictionary refinement - English to Indian languages
 11. What are the immediate tasks being undertaken?
 12. What do I gain by participating in collaborative activity?
    12.1 Hell and heaven
 13. Conclusions
 Acknowledgements
 References

(This writeup is also available as an HTML file.)

# 1. WHAT IS INFORMATION REVOLUTION?

The world is passing through a major revolution called the information revolution, in which information and knowledge is becoming available to people in unprecedented amounts wherever and whenever they need it. Those societies which fail to take advantage of the new technology will be left behind, just like in the industrial revolution.

The information revolution is based on two major technologies: computers and communication. These technologies have to be delivered in a COST EFFECTIVE manner, and in LANGUAGES accessible to people.

One way to deliver them in cost effective manner is to make suitable technology choices (discussed later), and to allow people to access through shared resources. This could be done throuch street corner shops (for computer usage, e-mail etc.), schools, community centers and local library centres.

# 2. WHAT ARE THE ADVANTAGES OF INFORMATION IN ELECTRONIC FORM?

## 2.1 COST

The major advantage of information in digital form (e.g., books in electronic form) is the lower cost of reproduction, storage and distribution. First, we take the reproduction cost.

For example, a compact disk (CD) can accommodate 500 books (of 500 pages of texts each with 300 words per page) at a CD reproduction cost of Rs.50/-. This works out to 10 paise per book. (The costs could be higher, if the book contains high-quality pictures as well. On the other hand, if newer technologies such as DVDs are used, the cost becomes lower.)

Internet also makes the distribution of digital books easier. Although, the bandwidth costs turn out to be higher, it has the advantage of the updated material available all the time. Thus, a judicious combination of CDs and web based material can be used with great cost effectiveness.

(Equipment needed to access electronic materials is expensive, but as we shall see shortly, the cost can be brought down. The cost reductions are truly dramatic if we are willing to share the use of equipment, and are willing to use equipment which is less "glamorous", but has all the important features. See below.)

## 2.2 CONVENIENT STORAGE

Storage of digital books is easier than paper as they take much less space. A CD containing 500 books (of text only) fits in a shirt pocket.

## 2.3 DISTRIBUTION EASIER

Distribution costs of digital books are also low. CDs can be distributed to libraries or a centre in each city, from where they can be accessed by the readers. The final access could be through a CD in a local machine or by using ordinary telephone lines to a nearby or remote centre having the digital book.

Essentially, the reproduction, storage and distribution cost of an electronic book is approaching zero.

## 2.4 SEARCH

A digital book can be used in new ways not possible with the paper copies of books. The computer can perform search in the digital library to help a user locate the information he wants rapidly. Powerful search engines are available that try to locate the information wanted by a user rapidly.

## 2.5 HYPER-TEXT

Digital books are increasingly being based on hyper-text which present the topics in a book, in different ways to the reader according to his requirements. Thus, the digital books need not be in linear format like the paper books.

## 2.6 EXPERT SYSTEM

Computer software (called expert systems) will be able to use the information in digital libraries to perform useful tasks for us inthe future.

There are many issues in the above. However, the following are taken up in detail below. (Other issues are not taken up here.)

(i). Hyper-text
(ii). Cost of equipment or computers to read the book.
(iii). Creation of content (author's effort and intellectual property rights)

## 3. WHAT IS HYPER-TEXT ?

A digital book can be so organized that it can be read in different ways based on topics, levels of detail needed, etc. For this the author has to connect relevant bite-sized chunks such as paragraphs, sections, etc. together, anticipating different uses. Links which connect different chnnks together, are called hyper-links. Thus, the book is not just a linear sequence of pages, but one in which there are many connections. This is called the Hyper text.

A computer allows the readers to follow different links of their choosing in the hyper text easily. For example, if you are reading this document in hyper-text form, you can jump across questions, get a list of questions, etc. More hyper links can be added to it depending on the needs of different readers.

# 4. WHERE ARE INDIAN LANGUAGES IN THE INFORMATION REVOLUTION?

We assume that for a real impact of the information revolution, it is important that the information is available in Indian languages. This requires that certain key technologies and resources should be available in Indian languages. Where are we with respect to these?

## 4.1 DIGITAL CONTENT

A key need is for the digital content (or electronic texts etc.) to be available in Indian languages, tailored to our environment and needs. Presently, very small amount of educational and informative material is available. Most web sites with Indian language content at present store newspapers. The information is of value for a short duration only. Compact disks (CDs) with Indian language content are few, and expensively priced.

## 4.2 MACHINE TRANSLATION

Automatic translation can allow a user to read electronic material from one language in another. Using the current state-of-the-art technology, it is not possible to build fully-automatic, general-purpose, high-quality machine translation system for any pair of languagesof the world.

However, it is possible to build language-access systems which allow a reader to read and understand electronic material in another language, provided he is willing to put in some effort. The output of such systems would not qualify to be called as a finished translation, and might also entail some training-to-read on part of the user.

Anusaaraka language access systems have been shown to be feasible from one Indian language to another. (Presently alpha-versions of the anusaaraka systems allow a Hindi reader to read material from five languages namely, Telugu, Kannada, Bengali, Marathi, and Punjabi). It should also be possible to build such systems from English to Indian languages, but a large scale prototype is yet to be built.

## 4.3 SPEECH

Speech processing systems for English are now available commercially, and are likely to become extremely important. They allow the computer to "read out" a given text (text-to-speech), and the computer to understand a spoken utterance (speech-to-text). The latter task is harder, but commercial systems for English have now appeared. Indian language speech technology is nowhere near. Accelerated effort needs to be made if we want to catch up in the next 5 years.

## 4.4 OPTICAL CHARACTER RECOGNITION

Optical character recognition is another technology that allows the computer to "read" a printed page. A printed page on paper can be scanned using a scanner, and its image (like a photograph) can easily be stored in the computer. However, for advantage to be taken regarding search and other language processing tasks (as discussed elsewhere in this writeup - digital content), characters have to be recognized from the image, and the image file is to be converted into a text file. These systems are available and are in routine use for English for past several years. Practical Indian language OCRs seem far away.

## 4.5 SEARCH AND KNOWLEDGE ORGANIZATION

An important area is that of searching the repository of digital content and knowledge. This allows readers to locate and access the information they need. Development of search technology for Indian languages requires: development of dictionaries, thesauri, subject index, document classification algorithms, etc. This technology is developing rapidly for English. The next logical step in this direction is the development of elaborate knowledge bases.

## 4.6 OTHER APPLICATIONS

There are a host of other applications based on NLP/Speech and related technologies. For example, font support in various applications, spelling correctors, reading aids, writing aids, summarisation aids, speaker identification, reading machine for the blind, etc.

## 5. RELEVANT TECHNOLOGIES

Here is a list of technologies that are needed in developing systems for the applications mentioned above.

1. Grammar Based Processing
    1.1 Word Analysers
    1.2 Sentence Analysers
2. Statistical Processing for NLP
3. Dictionary Building
    3.1 Among Indian Languages
    3.2 Between English and Indian Languages
4. Thesauri, Lexical Resources, and Knowledge-Base (KB)
    4.1 For Indian Languages
    4.2 Dealing with English
    4.3 Applications specific
5. Inference Methods
6. User Interfaces
7. Speech
    7.1 Signal Processing
    7.2 Acoustic-Phonetic Labelled Database Creation
    7.3 Statistical Processing & Pattern Recognition
    7.4 Grammar-based Processing

    7.5 Vocal Tract Modeling
    7.6 Compression
  8. OCR related technologies

## 5.1 BASIC RESEARCH

Basic research needs to be conducted in each of the above areas, in particular, grammar based processing, inference, organization of knowledge, etc. A special theme to be pursued is to relate Western theories of language with the Indian traditional theories of language. Areas of special promise are: vyakarana, navya-nyaya, and mimamsa. However, it is important to keep the basic research linked to practical applications.

## 6. ARE COMPUTERS AFFORDABLE?

The cost of computers or equipment to access digital books is high. A typical low cost computer which can operate independently costs around Rs.30,000/- in hardware and an additional Rs.10,000/- for some basic software (for email, word processing, spreadsheet, with say, Windows-98.) It is also difficult to share, as one user can tamper with the work done by another.

The key to reduce the cost is to :
1. Use innovative hardware-software combination.
2. Share the machine with several users (over time).

## 6.1 INNOVATIVE SOLUTIONS

Innovative solutions are available to lower the cost. These reduce the cost dramatically, if a number of "computer seats" are to be provided in a room or a building. The hardware cost of an independent computer (PC) would be around Rs.30,000/-. To this computer, one could connect, say, 8 thin clients such as Indian script GIST terminals at a price of Rs.11,000/- each. The total cost (Rs.1,18,000) divided by 9 "computer seats" (8 GIST terminals and one PC) turns out to be Rs.13,000/- per computer seat. (The cost can be reduced further to say Rs.8,000/- if the thin clients are manufactured on a large scale).

Free-software such as Linux and GNU (as opposed to Windows-98) can be used to support the above configuration. (Windows-98 would cost money, besides it does not have the capability to support such a configuration.)

It is important to realize the quality and power of free software. It not only includes software for internet, email, word processing, spread sheet but also includes a host of programming language compilers, software tools, database management, etc. Price of buying such software for Windows-98 would run into a lakh of Rupees. Not only this, the free software is usually faster and uses less memory space than the priced versions of such software for Windows-98.
Most important of all, source code is "open" and can be tinkered with, to suit our requirements.

With GIST terminals, all display and keyboarding capability in all Indian languages, comes as part of the hardware itself. As we have seen earlier, these reduce the cost per seat substantially. The limitation is that these are text-only seats. However, there would be one graphics seat in the above configuration namely the console on the PC.

Some amount of learning will also be needed because people are not familiar with (the interfaces of) the existing free software.

**6.2 SHARING THE USE OF COMPUTERS**

The above solution (rather than Windows-98) is also designed for shared use. It allows the same CD on a CD drive to be shared among several readers at the same time. With a single dialup phone connection, internet and email becomes available on every seat. This lowers the phone connect charges substantially. It also allows users to save their work on disk which is protected from other users modifying it (or even reading it, if so decided). There is no such protection on Windows-98. Thus, a computer seat can be shared among different users over time without the fear of their affecting each others work. Because of such rotection, these machines are also relatively immune from computer viruses.

If we assume the cost of a computer seat as Rs.13,000/-, the annual recurring expenses (maintenance 10%, interest on loan 15%, depreciation 25%) turn out to be Rs.6,500/- per year. This works out to Rs.550/- per month or Rs.20/- per day. If the machines can be put to use for 10 hours a day (27 days a month), the cost is Rs.2/- per hour.

**6.3 FUTURE COST PROJECTIONS**

Some believe that it is possible to reduce the retail cost of a terminal to Rs.4000. by clever engineering in the immediate future. (Or if the user is willing to use his TV as a display device, and a low cost keyboard, the cost can go down to below Rs.2000.)

**7. WHAT IS THE STATUS OF INDIAN LANGUAGE CONTENT IN ELECTRONIC FORM?**

There is an acute paucity of material in Indian languages in the electronic form. CDs available are few, and the internet also has very little material. If our people have to take benefits of the information society and India as a nation has to prosper, it is extremely important that knowledge becomes available in Indian languages in the electronic form.

# 8. HOW CAN WE CREATE ELECTRONIC CONTENT IN INDIAN LANGUAGES ?

The first thing that is required is that we should start creating content in our respective subject areas in Indian languages in electronic form.

If we choose the free-software model for electronic content, the whole process can be speeded up. In this model, we contribute our labour creating digital content. At the simplest level people can enter in the computer, existing published material whose copyright has expired or after obtaining permission for entering it. People can also create a new book as a free-text on the internet, and give permission to others to not only read it freely, but also modify it and place the new versions as free-text (with the same conditions). This permits and encourages thousands of readers to contribute their labour. One reader might help in proof-reading, another might build hyper-links, yet another might refine the material further. Editorial work would also be needed to control and coordinate the refinements being done by myriads of people, but that could again be distributed in a similar way albeit a little more carefully. A task that is highly suitable for creation through such work are mono-lingual and bilingual dictionaries, thesauri, etc.

The presence of software tools such as anusaaraka (a kind of machine translation software) that allow access across Indian languages can also help in making content in one Indian language available in another Indian language. The output from such tools can be further post-edited by people to produce translations and kept over the net.

# 9. CAN THE COLLABORATIVE METHOD FOR CREATION OF DIGITAL CONTENT WORK?

It is said that if everybody were to write parts of a book, there would be no continuity, besides repetitions and other problems in the book. The quality of books would be low. The answer turns out to be different from what one might expect to happen.

Let us take the case of software writing. In the case of software, a single mistake in any one of the modules can render the whole software totally unusable. However, in spite of such sensitivity to mistakes, free software continues to be written by myriads of programmers working together, and the greatest surprise is that it has turned out to be of higher quality than software written under companies under tight control. How has this come about? The number of people willing to read and test computer programs is very large, as a result errors in such open software get caught and corrected much faster than those written by companies. Of course, this requires a degree of coordination and control, which is brought about by the community itself without a tight central control. A single person coordinates, but he gets free-helping hands, and most importantly, when the coordinator wants to move on to other things, his coordinated software is taken over by another and continues to live.

If software can be written by such a multitude of authors, certainly books and texts can be written this way, and they are likely to be superior to other modes of their creation. They will also tend to get refined and updated even after the initial author(s) has moved on to other things.

To create digital content in our languages, the fastest way to move forward and involve thousands of people in its creation is to adopt the cooperative-collaborative model. No company can match the speed (and ultimately quality) in this model.

## 10. WHAT CAN I DO? WHAT ARE SOME OF THE CONCRETE TASKS THAT CAN BE DONE COLLABORATIVELY?

Digital content can be developed if people are willing to donate their labour ("shrama-daana"). Some example tasks are given below. The simplest task is that of data entry for existing high-quality material whose copyright has expired, or whose permission has been taken from the copyright holder. Similarly there are tasks such as proof-reading, converting existing content into hyper-text format, refining existing digital material, etc.

Preparation of exhaustive dictionaries and other lexical resources are needed for building machine translation and other systems for our languages. These should be taken up immediately, as they have a multiplier effect on the digital content (across languages).

### 10.1 DICTIONARY REFINEMENT

As a first task, the following six dictionaries are being placed on the internet for free download and refinement: Telugu to Hindi, Kannada to Hindi, Marathi to Hindi, Punjabi to Hindi, Bengali to Hindi. You can go through them, use it for your own purpose, as well as send us feedback about it for its correction, refinement etc.

### 10.2 DICTIONARY REFINEMENT - ENGLISH TO INDIAN LANGUAGES

As another concrete activity, dictionaries from English to Hindi, and to Telugu and Marathi are being created using the collaborative model. If you wish to participate in the task, contact the Akshar Bharati group at sangal@iiithyd.ernet.in.

## 11. WHAT ARE THE IMMEDIATE TASKS BEING UNDERTAKEN?

Creation and refinement of on-line and open dictionaries is being undertaken. (See the previous question.) You can participate in it from your place of work or home if you have access to computers and email. Contact us at the address given at the end.

## 12. WHAT DO I GAIN BY PARTICIPATING IN COLLABORATIVE ACTIVITY?

By participating in this collaborative activity and creating "free" content, you gain in several ways. First and foremost, you become part of a process which helps in preparing the nation and our languages for the next millenium. Second, if you make excellent contibutions you become known and get recognized by the community. Third, you acquire skills by participating in this activity, which will help you become entrepreneur, free lance person, or get jobs.

### 12.1 HELL AND HEAVEN

You might have heard of a story on heaven and hell. A person was taken to a visit to hell. It was lunch time, and much to his surprise, excellent food was being served. But soon there was much noise and commotion, and most of the food was spilled, with most people going hungry. The reason was that on everybody's hands were tied long spoons of several feet. As a result, the people could not eat without tossing the food in their mouths.

He then visited heaven, where the meal had just ended. The food was the same, and the people seem to have just finished a quiet and fulfilling meal. To his utter surprise, here too long sppons were tied to the hands of people. On enquiring, he learnt that the only difference was that here people had fed each other rather than trying to eat by themselves.

Collaborative creation of digital content is a similar activity. Either we can all starve and be frstrated, or we all work together in creating new resources to be used by everybody.

## 13. CONCLUSIONS

In this lecture, we have pointed out that for the information revolution to touch our society, it is important that information and knowledge is available to people at an AFFORDABLE COST, and in INDIAN LANGUAGES. This is crucial to the development and survival of our culture and society.

This requires at the digital content front, that a massive effort is undertaken for the creation of electronic content, that is suitable to our needs. These means creation of new content, as well as putting existing paper-content in electronic form. Common people as well as writers, teachers, and other professionals can contribute to this effort.

At the technology front, we need to develop systems for multi-lingual access, speech processing, optical character recognition, and search etc. for our languages. The development of these systems requires language resources, such as electronic dictionaries, and other lexical databases etc. to be built-up for Indian languages. Non-experts can also contribute to this task. The most suitable model for this is the voluntary-collaborative model or "free" software model.

**ACKNOWLEDGEMENT**

We acknowledge the support of Satyam Computers in activities pertaining to development of Indian language related technology. Anusaarakas from five Indian languages to Hindi were developed by IIT Kanpur and University of Hyderabad with the financial support of Dept. of Electronics, Govt. of India.

**REFERENCES**

Bharati, Akshar and Vineet Chaitanya and Rajeev Sangal, Natural Language Processing: A Paninian Perspective, Prentice-Hall of India, New Delhi, 1995,

Bharati, Akshar, Vineet Chaitanya, Amba P. Kulkarni, Rajeev Sangal, G. Umamaheshwar Rao, Anusaaraka: Overcoming the Language Barrier in India, To appear in "Anuvad: Approaches to Translation", Rukmini Bhaya Nair, (editor), Katha, New Delhi, 2000.

Narayana, V. N, Anusarak: A Device to Overcome the Language Barrier, PhD thesis, Dept. of CSE, IIT Kanpur, January 1994.